# High-level robot programming based on CAD: dealing with unpredictable environments

[1]Pedro Neto, [1]Nuno Mendes, [1]Ricardo Araújo, [1]J. Norberto Pires
[2]A. Paulo Moreira

[1]Department of Mechanical Engineering (CEMUC), University of Coimbra, Coimbra, Portugal
[2]Institute for Systems and Computer Engineering of Porto (INESC-Porto), Porto, Portugal

**Abstract:**

**Purpose** – The purpose of this paper is to present a CAD-based human-robot interface that allows non-expert users to teach a robot in a manner similar to that used by humans to teach each other. Another important issue addressed here has to do with how robots deal with uncertainty.

**Design/methodology/approach** – Intuitive robot programming is achieved by using CAD drawings to generate robot programs off-line. Sensory feedback allows minimization of the effects of uncertainty, providing information to adjust the robot paths during robot operation.

**Findings** – It was found that it is possible to generate a robot program from a common CAD drawing and run it without any major concerns about calibration or CAD model accuracy.

**Research limitations/implications** – A limitation of the proposed system has to do with the fact that it was designed to be used for particular technological applications.

**Practical implications** – Since today most of the manufacturing companies have CAD packages in their facilities, CAD-based robot programming may be a good option to program robots without the need for skilled robot programmers.

**Originality/value** – A new CAD-based robot programming system is proposed. Robot programs are directly generated from a CAD drawing "running" on a commonly available 3D CAD package and not from a commercial computer aided robotics software, making it a simple CAD integrated solution. This is a low-cost and low-setup time system where no advanced robot programming skills are required to operate it.

# 1. Introduction

## 1.1. Motivation

Increasingly, companies are changing and reinventing their production systems. Traditional manufacturing systems (often based on fixed automation and manual work) are being replaced by flexible and intelligent manufacturing systems, enabling companies to continue to be competitive in the global market (Kopacek, 1999). This competitiveness is reflected in the companies' capacity to respond/react quickly to market demands, producing more and better quality products at competitive prices.

Owing to its flexibility, programmability and efficiency, industrial robots are seen as a fundamental element of modern flexible manufacturing systems. Nevertheless, there are still some problems that hinder the utilization of robots in industry, especially in small and medium-sized enterprises (SMEs). SMEs have difficulty finding skilled workers capable of operating with robots. Therefore, new and more intuitive ways for people to interact with robots are required to make robot programming more accessible, easier and faster. The goal is that the instructor can teach a robot in a manner similar to that used by humans to teach each other, for example using CAD drawings, gestures or through verbal explanation (Neto *et al*., 2010a).

## 1.2. Objectives

Robot programming through the typical teaching method (using the teach pendant) is a tedious and time-consuming task that requires technical expertise. The goal is to develop methodologies that help users to program a robot in an intuitive way, quickly, with a high-level of abstraction from the robot specific language, and, if possible, without speeding too much money.

In this paper, a CAD-based system to program a robot from a 3D CAD drawing, allowing users with basic skills in CAD and robot programming to generate robot programs off-line, is presented. In addition, the 3D CAD package, Autodesk Inventor, that interfaces with the user is a well-known generic CAD package, widespread on the market at a relative low-cost. Starting from the CAD model of the robotic cell in study, the way the user generates a robot program is as simple as "drawing" the desired robot paths in the CAD environment. Later, the information needed is automatically extracted from the CAD environment, analysed and converted into robot programs. Note that the robot programs are not extracted neither from a computer aided manufacturing (CAM) software nor from a computer aided robotics (CAR) software. It means that we are proposing a simple CAD integrated solution for the robotics field.

Commercial CAR packages are powerful tools, which enable modelling, simulation and robot programming. Nevertheless, they have some disadvantages that hinder their use in companies, especially in SMEs. By comparing commercial CAR packages with a CAD-based robot programming system similar to that presented in this paper (Neto *et al.*, 2010b), it was found that the CAD-based system has some relative advantages:

• Low-cost. Since the construction of CAD models and the robot programming task are performed in the same environment/platform (Autodesk Inventor) the programming task becomes easier and cheaper;
• Short learning curve;
• Simplicity of use. The most time consuming task, the construction of the CAD model, is present in both systems.

CAD-based robot programming approaches work well if the environment of the robot tasks is well defined. In the other hand, there are situations which are likely to create errors or impede the normal operation of the robot:

• The CAD models do not reproduce correctly the geometry of the real scenario;
• Inaccuracies created in the robot calibration process;
• Inefficient fixtures that do not ensure the static character of the workpieces;
• A "foreign" object is introduced in the real environment.

In these cases, we can say that we are in the presence of a dynamic and unpredictable environment.

To perform successful manipulation robots depend on precise information about objects in their surrounding. In an unpredictable environment, such information cannot be given to the robot *a priori*, robots have to autonomously and continuously acquire information about their surrounding environment to support their decision making and react to unanticipated events. Sensory feedback allows a robot to recognize your work environment for itself, for example producing corrections (on-line) in pre-programmed robot paths (Figure 1). In fact, the integration of sensors into robotic platforms reduces the setup time, the need for accurate robot trajectory programming and promotes flexibility and the autonomous behaviour of robotic systems (Bolmsjö and Olsson, 2005; Johansson *et al.*, 2004).

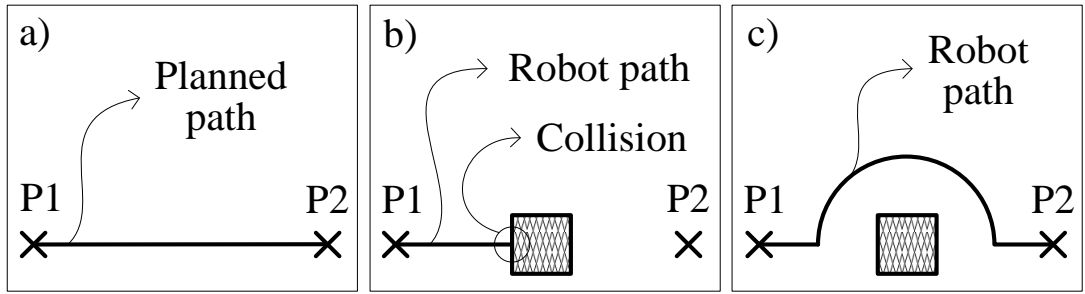

**Figure 1** (a) – planned path for a specific environment; (b) – a "foreign" object is introduced into the environment and collision occurs; (c) – sensory feedback is introduced, helping the robot to deal with the unpredictable environment (robot path is adjusted)

We validate our methods with two different real-world experiments for two different tasks, seam tracking and for applications that require the robot to follow a geometric profile while maintaining a contact force.

## 2. Related Work

In recent years, CAD technology has become economically attractive and easy to work with so that today millions of SMEs worldwide are using it to design and model their products. Already in the 80's, CAD was seen as a technology that could help in the development of robotics (Bhanu, 1987). Since then, a variety of research has been conducted in the field of CAD-based robot planning and programming.

Pires *et al.* (2004) proposes to extract robot motion information from a CAD DXF file and converting it into robot commands for welding purposes. A review of CAD-based robot path planning for spray painting is presented by Chen *et al.* (2009). Another study presents a method to generate 3D robot working paths for a robotic adhesive spray system for shoe outsoles and uppers (Kim, 2004). Nagata *et al.* (2007) proposes a robotic sanding platform where the robot paths are generated by CAD/CAM software. An example of a novel process that benefits from the robots and CAD versatility is the so-called incremental forming process of metal sheets (Schaefer and Schraft, 2005). Feng-yun and Tian-sheng, (2005) presents a robot path generator for the polishing process, where the cutter location data is generated from the postprocessor of a CAD system. As we have seen above, a variety of research has been done in the area of CAD-based robot planning and programming. However, none of the studies so far deals with a "global" solution for this problem.

Unpredictable environments pose a significant challenge because of their complexity and inherent uncertainty. Over the last few years, important studies have been carried out to deal with uncertainty in the robotics field: using models of "ideal" environments, sensory feedback, and implementing reasoning methods into robotic platforms (Bruyninckx *et al.*, 1991; Nayak and Ray, 1990). These concepts have evolved and recently, researchers have been successful in developing skills that can handle the complexity of dynamic and predictable environments (Kenney *et al.* 2009; Mendes *et al.*, 2010). A number of authors have devoted attention to sensor simulation, trying to mimic as closely as possible the behaviour of a real sensor, and thus integrating it (the virtual sensor) within a CAR platform (Cederberg *et al*., 2002; Brink *et al*., 1997; Bolmsjö and Olsson, 2005). Moreover, sensor information has been used to update robotic cell models in real-time, allowing to avoid problems such as collisions, kinematic singularities and exceeding of joint limits (Brink *et al*., 1997; Johansson *et al*., 2004).

The concept of seam tracking applied to robotic welding has been studied over the last two decades (Nayak and Ray, 1990). Recently, important work has been carried out in the integration of sensors to assist the robotic arc welding process (Fridenfalk and Bolmsjö, 2002; Bolmsjö and Olsson, 2005).

## 3. Robot Programming from CAD

Starting from a 3D CAD model of the robotic cell in study, the way the user generates a robot program can be as simple as "drawing" the desired robot paths in the CAD environment. Furthermore, to define the robot end-effector pose (position and orientation), it is necessary to know, not only the robot path positions but also the end-effector orientations in space. Therefore, after drawing the robot paths, simplified tool models should be placed along the paths. These models will define the orientation of the robot end-effector in each segment of the path (Figure 2).

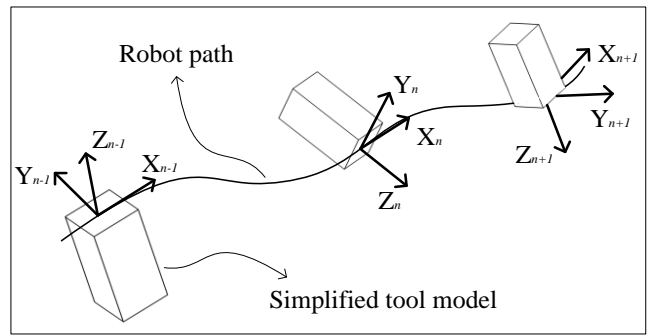

**Figure 2** simplified tool models defining the end-effector orientation

The information needed to program the robot will be extracted from the CAD environment by using an application programming interface (API) provided by Autodesk. This API allows the extraction of the points that characterize each of the different lines used to define a robot path; straight lines, splines and arcs. Moreover, the API also gives information about the transformation matrix of each *part model* represented in the CAD environment. The transformation matrix contains the rotation matrix and the position of the origin of the *part model* to which it refers, both in relation to the origin of the CAD *assembly model*. Later, the information extracted from the CAD is converted into robot programs (Video 1, 2010). A diagram with the procedure to extract 3D data from CAD and their conversion into a robot program is presented in Figure 3.

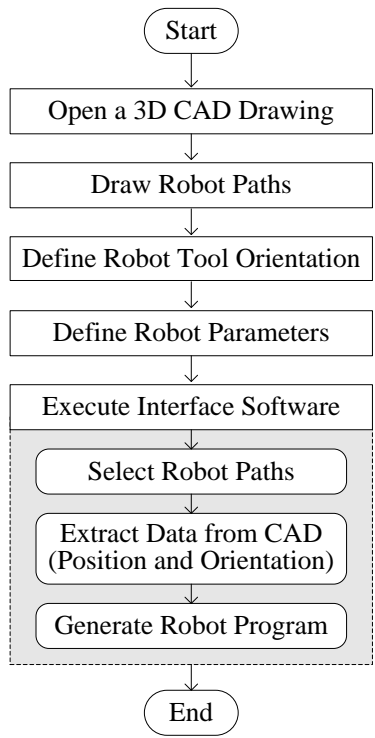

**Figure 3** extracting 3D data from CAD

### 3.1. Application programming interface

The Autodesk Inventor API shows the Inventor's functionalities in an object-oriented manner, allowing developers to interact with Autodesk Inventor using current programming languages; Visual Basic, Visual C#, Visual C++. In our proposed system, a standalone application was used to extract information from the CAD and the Autodesk Apprentice Server was used to display the CAD models on the screen, Figure 4. A flow chart, containing the method to automatically extract information about a straight line drawn in CAD, is shown in Figure 5.

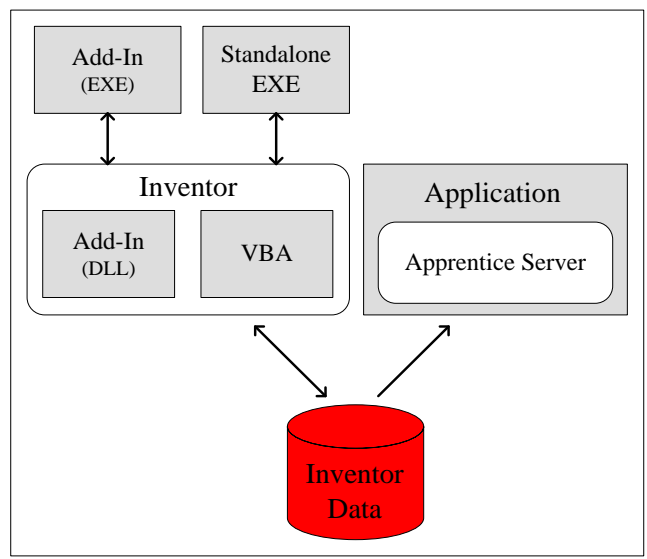

**Figure 4** accessing the Autodesk Inventor's API

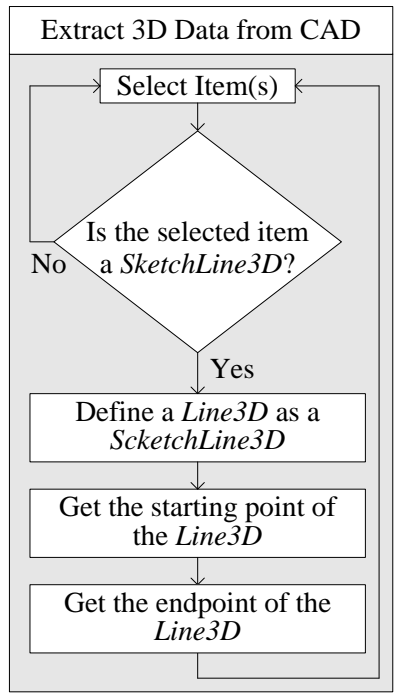

**Figure 5** extracting data from CAD (straight line)

### 3.2. Position and orientation in space

In order to off-line generate a robot program from a CAD environment and put it running in a real environment, the CAD cell should match with the real one. In other words, it is necessary to have all robot end-effector positions and orientations with respect to one or more reference frames known *a priori* by the robot. These frames are made known to the robot through a calibration process. Generally, this is a simple and non-time consuming process where the user needs to define the frame(s) within the CAD environment and then to teach the real robot about that frame(s)' pose in the real scenario (off-line to on-line mapping). When there are a significant number of frames to define, the calibration process can be lengthy and prone to error.

The API gives all the information (transformation matrices and path lines data) with respect to the origin of the CAD *assembly model*, the universe coordinate system {U}. Considering that a frame {B} is defined relative to {U} during the calibration process, from the API we have the transformation matrix of {B} relative to {U}, $^{U}_{B}T$. This means that frame {B} "makes the link" between the virtual and real world. Note that, as mentioned above, it is possible to define more than one frame if necessary, as the process is similar.

Since Autodesk Inventor considers the robot path lines drawn as a constituent of a single CAD *part model* (.ipt file) contained in the CAD *assembly model* (.iam file), the transformation matrix (relative to {U}) of that single *part model* defines the pose of the path lines. For the general case presented in Figure 6, the path line is part of the table top model in which the origin and orientation is defined by frame {E}. However, it is not necessary to know the orientation of the path lines as the API gives all the necessary points to define the path lines relative to {U}, for example the initial path point $^U P_{ini}$ (Figure 6). So it is necessary to achieve the path line points relative to frame {B}. In terms of establishing the robot end-effector orientation, frames {C} and {D} help to define the origin and orientation of simplified tool models in Figure 6. As mentioned, the API gives the transformation matrix of these models relative to {U}, $^U_C T$ and $^U_D T$. However, for robot programming purposes we wish to express frame {C} and {D} in terms of frame {B}, $^B_C T$ and $^B_D T$. For the case of $^B_C T$ we have:

$$^B_C T = {^B_U T} \cdot {^U_C T} \tag{1}$$

To find $^B_U T$, we must compute the rotation matrix that defines frame {U} relative to {B}, $^B_U R$, and the vector that locates the origin of frame {U} relative to {B}, $^B P_{Uorg}$. So, we know that:

$$^B_U T = \begin{bmatrix} ^B_U R & ^B P_{Uorg} \\ 0 \ 0 \ 0 & 1 \end{bmatrix} \tag{2}$$

Given the characteristics of a rotation matrix, $^B_U R = {^U_B R}^T$, and as we know $^U_B T$, the next step is to calculate $^B P_{Uorg}$. Considering a generic vector/point defined in {U}, $^U P$; if we wish to express this point in space in terms of frame {B} we must compute:

$$^B P = {^B_U R} \cdot {^U P} + {^B P_{Uorg}} \tag{3}$$

For the specific case of the initial path point in Figure 6, $P_{ini}$, since the API gives $^U P_{ini}$, from (3) we can write $P_{ini}$ relative to {B}:

$$^B P_{ini} = {^B_U R} \cdot {^U P_{ini}} + {^B P_{Uorg}} \tag{4}$$

Rewriting (3):

$$^B \left( ^U P_{Borg} \right) = {^B_U R} \cdot {^U P_{Borg}} + {^B P_{Uorg}} \tag{5}$$

The left side of (4) must be zero, so, from (4) we have:

$$^B P_{Uorg} = -{^B_U R} \cdot {^U P_{Borg}} = -{^U_B R}^T \cdot {^U P_{Borg}} \tag{6}$$

From (2) and (5) we can write:

$$_{U}^{B}T = \begin{bmatrix} _{B}^{U}R^{T} & -_{B}^{U}R^{T} \cdot {}^{U}P_{Borg} \\ 0 \quad 0 \quad 0 & 1 \end{bmatrix} \quad (7)$$

Now, we can rewrite (1) and achieve $_{C}^{B}T$. The same methodology can be used to achieve $_{D}^{B}T$ and any other transformation.

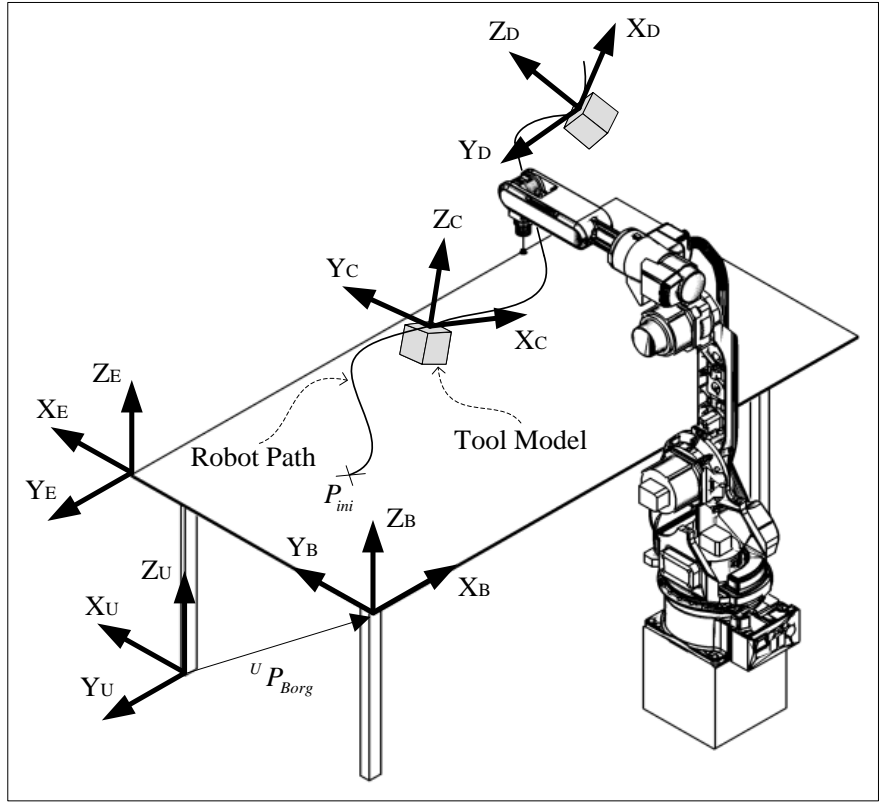

**Figure 6** system frames

### 3.3. Position and orientation interpolation

When an industrial robot is performing a pre-programmed movement and this one requires abrupt end-effector orientation changes, we must take special care because it can come into a situation where no one has total control over the end-effector orientation. This is particularly true when robot programs are generated off-line. The proposed solution to circumvent this problem is based on the implementation of linear smooth interpolation of end-effector positions and orientations (Feng-yun and Tian-sheng, 2005). The process involves the following steps:

- Identification of risk areas (paths). This is done by analyzing the CAD model and manually defining those areas in the drawing.
- Discretization of the risk path in equally spaced intervals.
- Calculation of end-effector orientations for each interpolated path point. The new path is smoother than the initial (Figure 7).

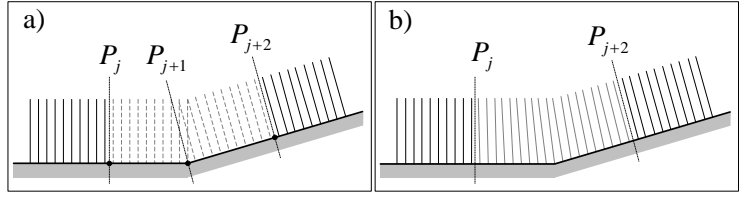

**Figure 7** (a) – end-effector pose before interpolation; (b) – end-effector pose after interpolation

Consider $r(k) = \begin{bmatrix} r_x(k) & r_y(k) & r_z(k) \end{bmatrix}^T$ a generic end-effector position generated at the discrete time $k$ and defined in $\begin{bmatrix} P_j & P_{j+2} \end{bmatrix}$, (Figure 7). $P_j$, $P_{j+1}$ and $P_{j+2}$ are known end-effector poses, extracted from the CAD drawing (see section 4.1.2). For the profile in Figure 7 (possible area of risk) we will separate the interpolation in two sections, $S_1$ and $S_2$; $S_1 \in \begin{bmatrix} P_j & P_{j+1} \end{bmatrix}$ and $S_2 \in \begin{bmatrix} P_{j+1} & P_{j+2} \end{bmatrix}$. The calculations are presented for section $S_1$ but for other sections the procedure is the same. So, $r(k)$ is calculated using both the known data points from CAD ($P_j$, $P_{j+1}$) and the profiling velocity $v(k)$:

$$v(k) = \begin{bmatrix} v_x(k) & v_y(k) & v_z(k) \end{bmatrix}^T \quad (8)$$

It is assumed that the magnitude of $v(k)$, $|v(k)|$, is a constant. Considering $r(k) \in \begin{bmatrix} P_j & P_{j+1} \end{bmatrix}$, a direction vector $W$ can be defined as:

$$W = P_{j+1} - P_j \quad (9)$$

From (8) and (9), each directional velocity profile is obtained by:

$$v_i(k) = |v(k)| \cdot \frac{W_i}{|W|}, \quad (i = x, y, z) \quad (10)$$

From (10), using a sampling width $\Delta t$, the interpolated position $r(k)$ is given by:

$$r(0) = P_j^T = \begin{bmatrix} P_{j,x} & P_{j,y} & P_{j,z} \end{bmatrix} \quad (11)$$

$$r(n) = P_{j+1}^T = \begin{bmatrix} P_{j+1,x} & P_{j+1,y} & P_{j+1,z} \end{bmatrix} \quad (12)$$

$$r_i(k) = r_i(0) + v_i(k) \cdot k \cdot \Delta t, \quad \begin{cases} (i = x, y, z) \\ (k = 1, ..., n-1) \end{cases} \quad (13)$$

Note that $n$ represents the number of interpolated points.

A quaternion interpolation algorithm (spherical linear interpolation – Slerp) to interpolate smoothly a sequence of end-effector orientations was used. For the profile in Figure 7 we will interpolate end-effector orientations between $P_j$ and $P_{j+2}$. Given two known unit quaternions, $Q_0$ (from $P_j$) and $Q_n$ (from $P_{j+2}$), with parameter $k$ moving from 1 to n-1, the interpolated end-effector orientation $Q_k$ can be obtained as follows:

$$Q_k = \frac{\sin\left(\left(1-\frac{k-1}{n-1}\right)\cdot\theta\right)}{\sin\theta}\cdot Q_0 + \frac{\sin\left(\frac{k-1}{n-1}\cdot\theta\right)}{\sin\theta}\cdot Q_n, \quad k \in [1 \quad n-1] \tag{14}$$

Where:

$$\theta = \cos^{-1}(Q_0 \cdot Q_n) \tag{15}$$

### 3.4. Robot program generation

Using the information extracted from the CAD environment, the system presented here is able to generate robot programs for specific robotic applications. The code generation process is divided into two distinct phases:

- Definition and parameterization of robot positions/orientations, reference frames, tools, etc. The end-effector positions and orientations extracted from CAD are used to define the robot path target poses (16). When confronted with risk areas the interpolation algorithms automatically generate the appropriate end-effector poses for these areas. From (3) we have the end-effector positions $^BP$; from (1) the transformation matrix $\left(^B_C T\right)$ containing the rotation matrix, which in turn is used to calculate the end-effector orientation in the form of quaternions or Euler angles; from (13) the interpolated positions $r(k)$; and finally from (14) the interpolated orientations (quaternions) $Q_k$.

$$P = \underbrace{x, y, z}_{^BP \text{ and } r_i(k)}, \underbrace{q1, q2, q3, q4}_{^B_C T \text{ and } Q_k} \tag{16}$$

- Body of the program. A robot program contains predominantly robot motion instructions (linear, joint, circular or spline robot movement). These movement instructions are selected according to the type of lines used in the CAD drawing to define the robot paths.

## 4. Experiments

Two different experiments are discussed, and in both cases, robot programs are generated off-line from a CAD drawing. In the first experiment, seam tracking, robot paths are adjusted with the information received from a laser camera attached to the robot. In the second experiment, a robot follows a geometric profile while maintaining a contact force, robot paths are adjusted with the information received from a force/torque (F/T) sensor attached to the robot wrist.

To better visualize the robot path adjustments provided by sensory feedback, the robotic space was forced to become a more "viewable" unpredictable environment by purposely making a rough calibration process. Often, calibration errors arise from the little time and attention devoted to the robot calibration process.

### 4.1. Seam tracking

### 4.1.1. Experimental setup

The experimental setup of the robotic platform (Figure 8) is the following:

- An industrial robot ABB IRB 2400 equipped with a S4C+/M2000 controller.
- A computer running Microsoft Windows Xp.
- A laser camera DIGI-I/S from Servo Robot.

The computer is running a CAD package (Autodesk Inventor) and the developed software interface, which receives data from CAD, interprets the data received and generates robot programs. The robot can be remotely controlled and managed by the software interface, which uses an ActiveX named PcRob for such purposes. The laser camera is connected with the robot controller *via* serial port.

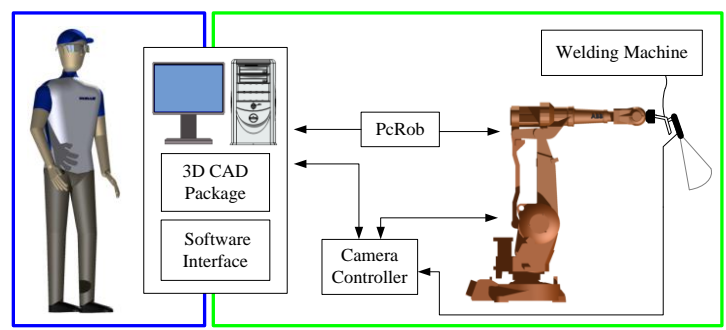
**Figure 8** system architecture

### 4.1.2. CAD model

The CAD *assembly model* from which a robot program will be generated does not need to accurately represent the real cell in all its aspects (Figure 9). On the contrary, it can be a simplified model containing the "important" information. As an example, the robot tool length, robot paths and relative positioning of CAD models should represent the real scenario, however, the models appearance do not need to be exactly equal to the real objects.

For this particular experiment, the CAD *assembly model* should contain the workpieces to be welded, the robot paths and the robot tools with the desired torch orientation for each path segment. In terms of risk areas, there is only one abrupt tool orientation change (Figure 9).

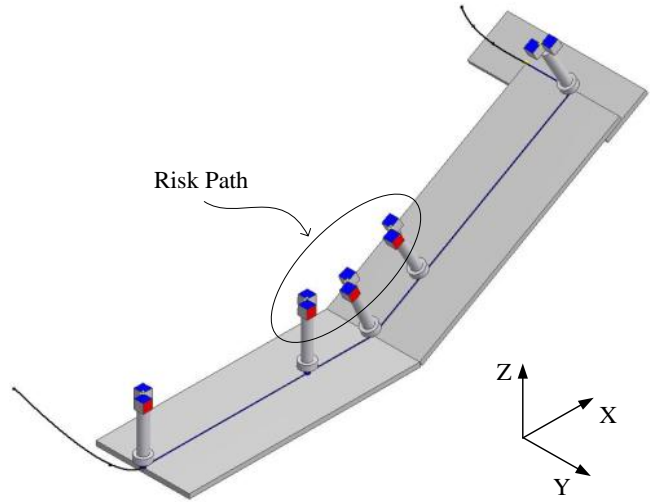

**Figure 9** CAD *assembly model* of the workpieces to be welded (butt joint). Note: a robot program will be generated from this model

### 4.1.3. Path adjustment

Analyzing the incoming data from the laser camera, the implemented control system decides which end-effector adjustments should be applied to the main paths extracted from CAD. The system *modus operandi* is relatively simple:

1  Definition/calibration of the robot tool to match with the robot reference frame.
2  The laser camera is configured with information about the welding joint and the desired vertical and/or horizontal distances (tool standoff) that the torch must maintain to the welding joint.
3  Features from the workpiece profile are extracted and matched against the predefined joint templates and tolerances.
4  The automatic end-effector adjustment is achieved by a closed loop position control that promotes compensation of the errors in Y and Z directions. Correction data are acquired with a sample rate of 5 Hz.

### 4.1.4. Results and discussion

Results showed that the CAD-based robot programming system is easy to use and within minutes an untrained user can generate a robot program for welding purposes. However, in the real scenario (Figure 10) we have a dynamic environment where robot path adjustments are required. Figure 11 shows the robot path adjustments/corrections (in the Y direction) made by the robot during the seam tracking process (Video 2, 2010).

As the robot only allows path adjustments at a frequency of 5 Hz, for higher welding speeds the path correction does not appear so smooth. Another limitation is the low robot resolution (0.01 mm), making the path adjustment process more abrupt.

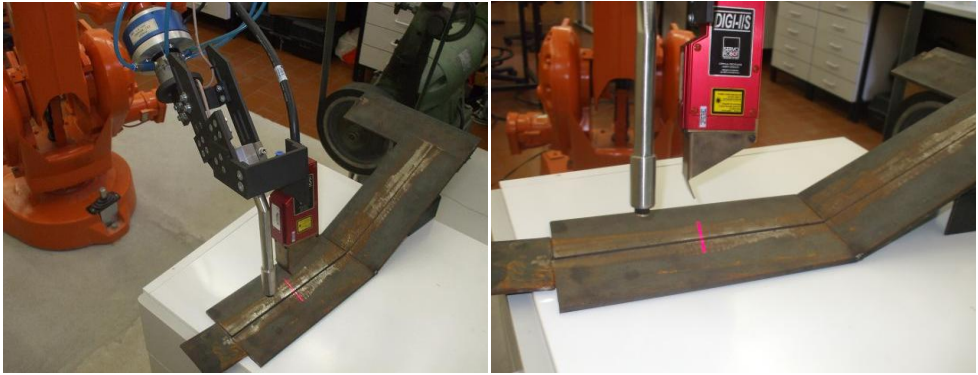

**Figure 10** robotic cell

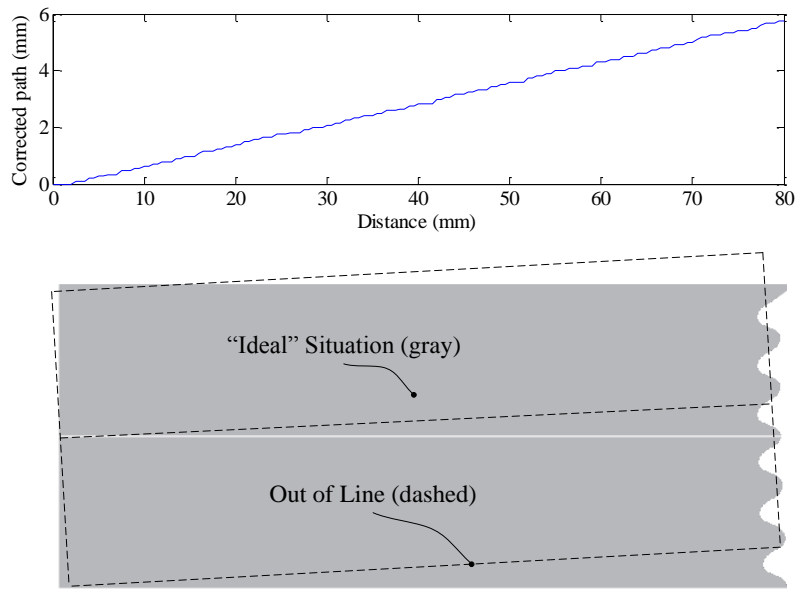

**Figure 11** path adjustments in Y direction (robot velocity 10 mm/s)

## 4.2. Profile following

### 4.2.1. Experimental setup and features

The experimental setup of the robotic platform (Figure 12) is the following:

- An industrial robot Motoman HP6 equipped with the NX100 controller.
- A computer running Microsoft Windows Xp.
- A six degrees of freedom (DOF) F/T sensor from JR3.
- A local area network (LAN), Ethernet and TCP/IP based, used for robot-computer communication (100 Mbps).

The computer is running Autodesk Inventor and the developed software interface. This interface generates robot programs from CAD and manages the force control system, acquiring data from the F/T sensor and sending motion commands (adjustments) to the robot. The software interface communicates with the robot using a software component named MotomanLib. The ActiveX component JR3PCI is used to acquire force and torque data from the F/T sensor. The robot pose is adjusted with a sample rate of 20 Hz.

As in the previous experiment, the robot program is generated from a CAD drawing (Figure 13). The real work environment is an unpredictable environment due to the "uncertainty" that comes from an inaccurate calibration process and due to the surface roughness of the workpiece. The robot tool should follow a geometric profile while maintaining a contact force. In order to facilitate the analysis of experimental results, a ball-shaped tool was mounted on the robot's end-effector.

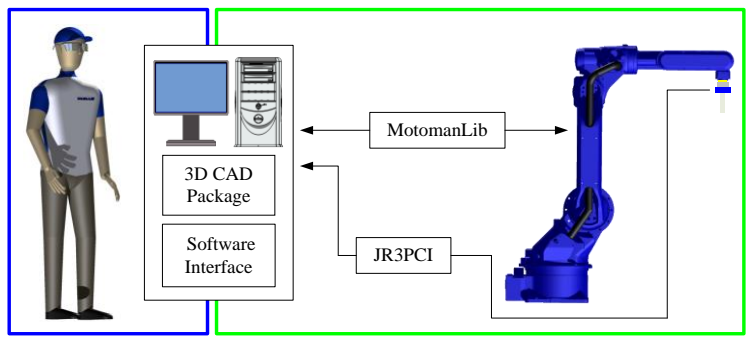

**Figure 12** system architecture

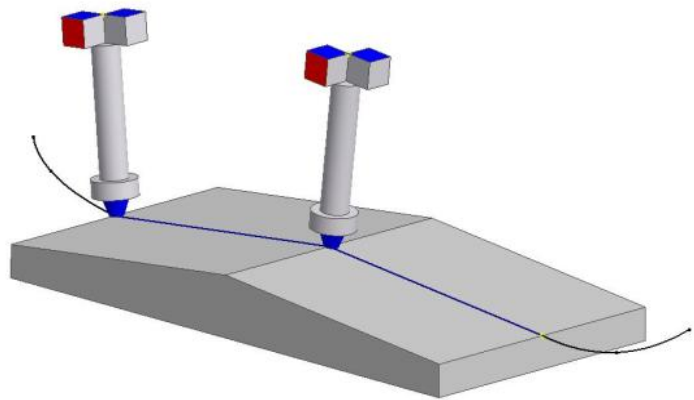

**Figure 13** CAD *assembly model* of the working profile.

### 4.2.2. Results and discussion

Regarding the generation of the robot program from a CAD drawing, this experiment showed similar results to those of section 4.1.4. From the incoming data from the F/T sensor, the implemented force and robot displacement control system (Fuzzy-PI and PI reasoning) decides which displacements should be applied to the robot end-effector to achieve satisfactory performance (Mendes *et al*., 2010; Video 3, 2010). The force control system ensures that the contact forces are maintained at a constant value, adjusting the pre-programmed robot paths extracted from CAD (Figure 14 and Figure 15). The graphs of Figure 14 show some force fluctuation due to the roughness of the surface and the noise of F/T data.

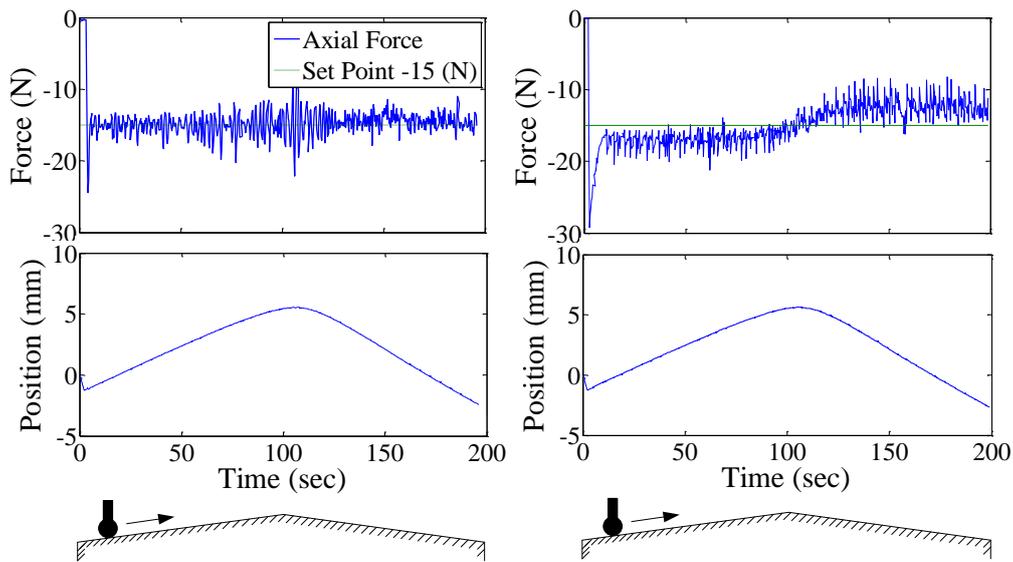

**Figure 14** experimental results by using a Fuzzy-PI controller (at left) and PI controller (at right)

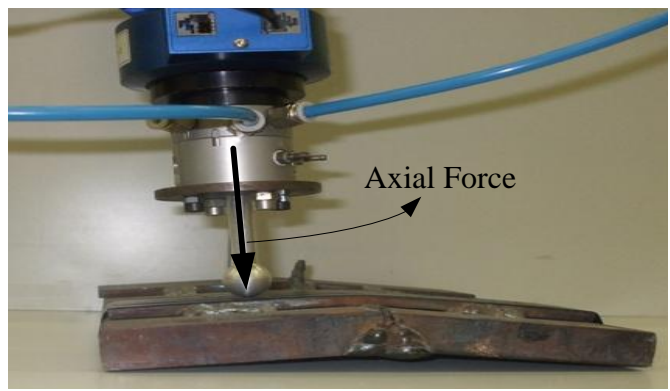

**Figure 15** robot tool in contact with the real workpiece

### 4.3. Overall results

Some problems can occur when external sensors are used to on-line adjust robot motion:

- Collisions between the external sensor and the surrounding workspace;
- Situations in which the robot arm is sent to a location outside of the robot working area;
- Kinematic singularities;
- Poor choice of process parameters;
- The communications delay between the external sensor and the robot controller can produce a negative effect on the proper definition of the robotic task.

In order to avoid the above mentioned problems the operator should ensure that the workpieces are inside the working area of the robot, no collisions occur and kinematic singularities are identified.

During effective robot operation, if a failure or malfunctioning is detected, two different situations can be considered: task abortion or activation of a reactive task. After aborting the process, the restarting of the system can be a complicated issue, depending on the type of robotic task. For example, for an arc welding application, restarting the system requires at least placing the torch at the point where the robot stopped.

## 5. Conclusion and future work

A CAD-based robot programming system with capacity to deal with dynamic and unpredictable environments was presented. Results showed that the proposed platform opens new possibilities for intuitive robot programming. It means that an untrained operator can generate a robot program for a specific task within minutes. Moreover, since the construction of the CAD models and robot programming task are performed in the same platform the entire robot programming process becomes easier and cheaper. This is very important for SMEs that produce small batches of products and need to constantly reprogram the robotic cells. In addition, sensory feedback enables the robot to be more flexible when confronted with product changeover. By adding sensory feedback to the robotic platforms we ensure that the robot manoeuvres in an unpredictable environment, damping possible impacts and increasing the tolerance to positioning errors from the calibration process or from the construction of the CAD models.

Future work will be required to proceed with the development of methodologies which would facilitate sensor integration in robotic platforms, especially for when robots are programmed off-line.